# CLASSIFICATION OF VIRAL PNEUMONIA X-RAY IMAGES WITH THE AUCMEDI FRAMEWORK

*Pia Schneider[1,2], Dominik Müller[1,2], Frank Kramer[1]*

[1] IT-Infrastructure for Translational Medical Research, University of Augsburg, Germany
[2] Medical Data Integration Center, University Hospital Augsburg, Germany

## ABSTRACT

In this work we use the AUCMEDI-Framework to train a deep neural network to classify chest X-ray images as either normal or viral pneumonia. Stratified k-fold cross-validation with k=3 is used to generate the validation-set and 15% of the data are set aside for the evaluation of the models of the different folds and ensembles each. A random-forest ensemble as well as a Soft-Majority-Vote ensemble are built from the predictions of the different folds. Evaluation metrics (Classification-Report, macro f1-scores, Confusion-Matrices, ROC-Curves) of the individual folds and the ensembles show that the classifier works well. Finally Grad-CAM and LIME explainable artificial intelligence (XAI) algorithms are applied to visualize the image features that are most important for the prediction. For Grad-CAM the heatmaps of the three folds are furthermore averaged for all images in order to calculate a mean XAI-heatmap. As the heatmaps of the different folds for most images differ only slightly this averaging procedure works well. However, only medical professionals can evaluate the quality of the features marked by the XAI. A comparison of the evaluation metrics with metrics of standard procedures such as PCR would also be important. Further limitations are discussed.

## 1. INTRODUCTION

Viral pneumonia is pneumonia that is caused by a virus. It causes inflammation in one or both of the lungs (1). Respiratory viruses can be detected with conventional virus diagnostic methods such as culture, antigen detection or serological assays and the newer PCR-based methods. The PCR-based methods are two to five times more sensitive but remain unpleasant for the patient, especially if the specimens are obtained from the lower-respiratory tract. Therefore the American Thoracic Society recommends, that the diagnosis of pneumonia should be made based on chest radiography (2).

To aid medical professionals in their decision-making Machine Learning combined with Explainable Artificial Intelligence (XAI) might be useful. We use the AUCMEDI-Framework (3) to train a deep neural network to classify chest X-ray images as either "normal" or "viral pneumonia". The most important parts of each image for the classification are visualized with XAI. This way medical doctors could validate the results.

## 2. METHODS

### 2.1 Overview over the Pipeline

After the dataset is split in training-, test- and validation-sets (section 2.3) the Neural Network is setup and trained with k-fold cross-validation (section 2.4). Afterwards two ensembles of the different cross-validation folds are build (section 2.6), one with Random-Forests and one Soft Majority Vote Ensemble. Finally XAI is applied to the different folds (section 2.7).

### 2.2. Image Source

X-ray images from Kermany et al. (4) are used. The authors collected 5232 X-ray images of children between one and five years and provide them on kaggle (5). The chest X-ray images were acquired as part of the childrens' routine care and labeled by two expert physicians as "normal", "bacterial pneumonia" or "viral pneumonia". We use only the "normal" and "viral pneumonia" images.

### 2.3 Splitting the Dataset

The used dataset contains 2686 X-ray images. They are about equally distributed between the two classes "normal" and "viral pneumonia" (see figure 1).

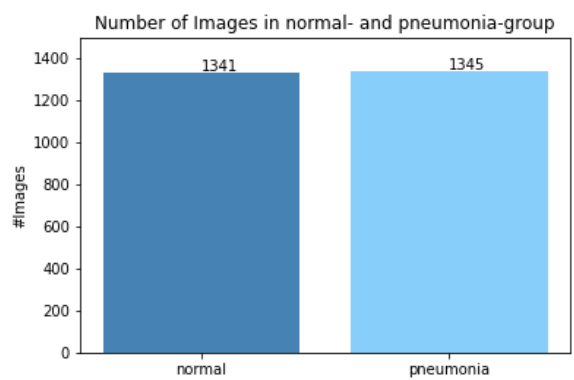

**Figure 1:** Distribution of the images in the different groups before splitting in train- and test-set.

This is advantageous as no class-weighting has to be done during training.

The dataset needs to be split into several sets: In general, at least a training-, a test- and a validation-set are needed. While the training-set is used for the actual training of the neural network, the test-set is a set that is never trained on. Therefore it can be used to evaluate the performance of the training. After each epoch of training, the training is validated. The validation-metric can be used for regularization by early stopping to avoid overfitting. That means, however, that the test-set can't be used for validation, because even if the model is not trained on the validation-set there is some information-leak, as the hyperparameters of the model are based of the model's performance on the validation-set. This can lead to the phenomenon that the model performs well on the validation-data, but much worse on new data. Therefore the model needs to be evaluated on independent test-data (6).

**Stratified k-fold Cross-Validation:** In this work stratified k-fold cross-validation (7) with k=3 is used to generate the validation-set. K-fold cross-validation means, that the training-set is randomly divided into k equally sized parts and training is performed k times. With *stratified* cross-validation the dataset is split in such a way that each fold contains approximately the same percentage of samples of each target class as the complete set (8). In each run, a different part serves as the validation-set while the other two parts serve as the training-set. An extreme form of k-fold cross-validation is leave-one-out-validation. There, training is performed n times, with n being the size of the dataset. Each time only one data-point (for example image) serves as validation-set. But this is not performed here, because it is very computationally expensive. Advantages of k-fold cross-validation are that it matters less how the data get divided and that the variance of the calculated evaluation metrics over all folds is reduced (9).

However, in order to calculate those metrics over all folds the predictions of the three individual folds need to be combined. This can be done with various ensemble methods, for example Random-Forests or Soft Majority Vote. In this work the different folds are combined with a Random-Forests Ensemble as well as with a Soft Majority Vote Ensemble (section 2.5). As Random-Forests are a Machine-Learning Algorithm that need to be evaluated themselves a different test-set is necessary for the evaluation of the Random-Forest-Ensemble. For this reason the complete set is in a first step divided into three sets: train (70%), test_models (15%) and test_ensemble (15%, see figure 2). In a second step the train-set is divided for k-fold cross-validation (k=3) into 3 folds of a test- and validation-set each (see figure 3).

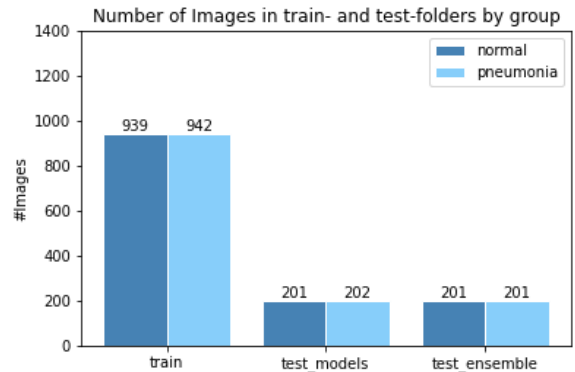

**Figure 2:** Number of images in the training-set (train), test-set for the models (test_models), which is used for evaluation of the best model of each fold and test-set for Random-Forest ensemble (test_ensemble).

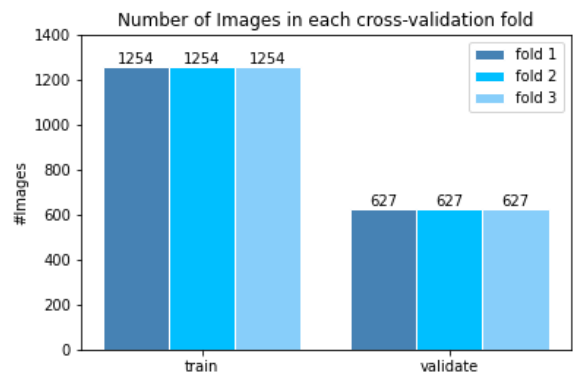

**Figure 3:** Number of images in each train- and validation-set for cross-validation fold.

### 2.4 The Neural Network

**Architecture:** MobileNetV2 (10,11) is chosen as architecture. Advantages of MobileNetV2 are that a version that is pretrained with ImageNet exists, which can be used for transfer-learning. Furthermore MobileNetV2 is relatively small, so that it can be applied on mobile devices or other hardware without much resources. This is useful, because this way prediction on individual images could be done in hospitals, which usually don't have large computational resources for machine learning.

**Activation-output:** The Softmax-function is used as the last-layer activation function. It ensures for every image each class is assigned a probability between 0 and 1 and the probabilities over classes for every image add up to 1 (12).

**Loss-function:** The loss function defines the quantity that a model should seek to minimize during training (13). Categorical cross-entropy is the appropriate loss-function for multiple class single label classification problems, therefore it is chosen here. It quantifies the difference between the labels and the predicted values (14,15).

**Image Augmentation:** Image-augmentation increases the image-set is artificially by adding small transformations to the original images such as rotations or changes of the contrast or saturation. It can be applied offline or online. With offline image-augmentation those transformations are applied to the images and the images are saved back on disk before they are loaded again for training. With online image-augmentation the transformations are applied to each image when loaded with the data generator (16). AUCMEDI supports both methods. However, this is not applied here, because it doesn't bring additional advantages in this particular case and is computationally expensive. It might be useful for more difficult classification-problems, or in cases when only few images are available for training.

**Callbacks:** The Callbacks EarlyStopping and ModelCheckpoint are used here. They help to reduce the risk of overfitting. The EarlyStopping Callback furthermore reduces unnecessary training-time.

**Transfer-learning:** The goal of transfer-learning is to use pretrained models on another classification task. Since the pretrained models were pretrained on a different and very large image-set (mostly ImageNet) their trained weights won't fit perfectly for the task at hand (the closer the better). The weights of the pretained models are most likely still better than an untrained model. Therefore the pretrained model can be used. The weights of the pretrained model of all layers but the classification head are then frozen during training of the new task. When the weights wouldn't be frozen they would be "deleted" by the first epoch of training. The classification-head, however, can't be frozen, because it needs to be adapted to the new task. After some epochs of training the freezing is undone, so that the weights can be adapted to the new task. With transfer-leaning models can be trained even when fewer data/images are available. Furthermore time is saved, because the weights are preinitialized.

**Metrics:** Metrics that are defined during the Neural Network definition are shown after each epoch of training (if verbose=1). In this case the appropriate – solely included – metric is categorical_accuracy. More metrics could be defined and shown, but this isn't done here as validation-loss is monitored for hyperparameter-tuning and for evaluation purposes metrics calculated on predictions of the test-set are more important.

## 2.5 Evaluation Metrics

Metrics for evaluation of the best models of each fold are calculated on the predictions test-set (15% of total data). The Confusion Matrix (17), Classification Report (18), and ROC-Curve (19) are calculated with functions of scikit-learn (20), after transformation of the data.

**Confusion Matrix:** The Confusion Matrix shows the number of True Positives (TP), False Positives (FP), True Negatives (TN) and False Negatives (FN) in a compact manner.

**Classification Report:** The Classification Report gives an overview over precision, recall and f1-score for the different classes as well as accuracy, macro average and weighted average precision, recall and f1-score (21) over all classes.

**Table 1:** Explanation of the metrics computed by a classification report with scikit-learn.

| Metric | Description | Formula |
|---|---|---|
| **Precision (pr)** | Ability of a classifier not to label a negative example as positive = 1 – False Discovery Rate | TP/(TP+FP) |
| **Recall (rec)** | Ability of a classifier to find all positive examples = True Positive Rate (TPR) | TP/(TP+FN) |
| **f1-score** | weighted average of precision and recall | 2 * (pr * rec) / (pr + rec) |
| **macro-f1-score** | class imbalances are not taken into account | |
| **weighted-f1-score** | weighted by support | |
| **accuracy** | fraction of correct predictions | |

The macro-f1-score is saved for later comparison, as the f1-score combines metrics in one score and class imbalances shouldn't be taken into account in medical classification, because often the interesting class is under-represented.

**ROC-Curve:** The ROC-(Receiver-Operating-Characteristic)-Curve (22) plots the False-Positive-Rate (FPR) against the True-Positive-Rate (TPR) for different thresholds of classification.

Example: If predictions for viral pneumonia for 10 images were [0.1, 0.9, 1.0, 0.5, 0.4, 0.6, 0.8, 0.2, 0.7], there *could* be 10 different thresholds for classifying these 10 images as viral pneumonia, namely 0.1, 0.2, 0.3, 0.4, 0.5, 0.6, 0.7, 0.8, 0.9 and 1.0. Depending on which of these images actually *are* viral pneumonia images the different threshold would lead to different False-Positive and False-Negative-Rates.

For example, if images with predictions [0.1, 0.2, 0.3, 0.4, 0.5] are normal and images with predictions [0.6, 0.7, 0.8, 0.9, 1.0] are viral pneumonia, then the False-Positive-Rate for Threshold >= 0.2 would be

$$FPR = FP / (FP + TN) = 4 / (4+1) = 0.8$$
$$TPR = TP / (TP + FN) = 5 / 5 = 1$$

But for Threshold >= 0.8 it would be

$$FPR = FP / (FP + TN) = 0$$
$$TPR = TP / (TP + FN) = 3 / (3+2) = 0.6$$

In the ROC-Curve the False-Positive-Rate and the True-Positive-Rate are plotted against each other for different thresholds. For point (1, 1) all "positives" were correctly identified, but also none of the "control" samples was correctly identified. Or, in other words, the True and the False Positive Rate is 1. The diagonal line shows where True Positive Rate = False Positive Rate. This would be realized by chance, if the predictor worked at random. The Area under the Curve (AUC) serves as quality-measurement. In the worst case (random performance) it is 0.5 and in the best case 1.0.

Here, additionally to the ROC-Curve for each fold, an average ROC-Curve of all three folds is calculated. The calculation isn't trivial, because for each fold different False- and True-Positive-Rates exists. For this reason, a mean can't be calculated because the FPR and TPR-vectors have different lengths and are differently distributed. For this reason, all FPRs are collected into one vector and the TPRs for these values are interpolated (23).

### 2.6 Ensemble Learning

With cross-validation three different models are produced, one for each fold. The predictions of these 3 models can be combined into an *ensemble* in order to produce one common - most likely better and more stable - prediction. Here, a Random-Forest- and a Soft-Majority-Vote-Ensemble is calculated.

**Random-Forest Ensemble:** Random forests (24) are an ensemble method themselves. They work the following:

1. Bootstrapped Dataset: Random sample from the dataset are drawn. One sample can be drawn several times. Therefore the bootstrapped dataset has the same size as the original dataset but doesn't contain all data.
2. Random Decision Trees: A decision tree with a random number of variables – here possible variables are fold1, fold2, fold3 – in each step is build based on the bootstrapped dataset. This is called feature bagging.
3. Back to Step 1 and repeat. This way many random decision trees are generated.
4. Tree bagging for the prediction of new data: The most common decision of the decision trees is the final decision.

**Soft Majority Vote Ensemble:** In order to create a Soft Majority Vote Ensemble predictions of the models are summed to create a new ensemble predictions matrix. For every picture the category that has the maximum value for this ensemble predictions-matrix is taken as the final prediction.

The predictions produced by the ensembles are evaluated with the same metrics as the individual folds (classification-report, confusion-matrix, ROC-curve).

### 2.7 Explainable Artificial Intelligence (XAI)

It is the goal of explainability (XAI) to show in human-readable format which features of the input where important for the deep-learning network. This might help for debugging, for example if it is shown that the network learned from background-information that are not important (for example different format of the pictures, or marking symbols etc.). Here, Grad-CAM (25) and LIME (26) are used as XAI-Algorithms.

**Grad-CAM:** With Grad-CAM the Output-Feature-Maps of the Convolutional-Layers are taken and every channel of this map is weighted with the gradient of the class that is predicted. Therefore it is shown (in the heatmap) how much the input-image activates the class.

**LIME:** LIME generates Superpixels. The Superpixels are generated by a segmentation of the image into different meaningful parts. These Superpixels are switched on and off and pictures are generated were some Superpixels are switched on and others of. The LIME-Algorithm calculates the influence of every Superpixel on the prediction of the class. The Superpixels with the greatest influence are shown.

All XAI-algorithms are calculated based on the predictions on one model. K-fold cross-validation produces k models (one for each fold). For this reason the heatmaps, which are produced by the Grad-CAM algorithm for each fold are averaged and then laid over the corresponding images.

## 3. RESULTS

### 3.1 Training History

The Training-History (see figure 4) shows that transfer-learning helps to increases the speed of learning.

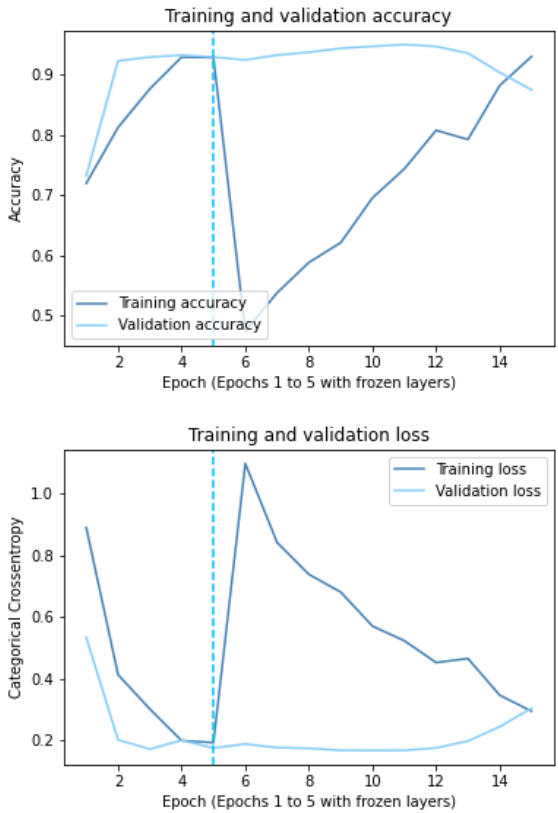

**Figure 4:** Training and validation accuracy (upper plot) and loss (lower plot) for fold1. The training histories for the other two folds are very similar.

After the layers are unfrozen after 5 epochs, there is a sharp drop in training accuracy and increase in training loss, because due to the unfreezing of the layers the correct weights have to be found again. For the validation loss and accuracy this abrupt chance after epoch 5 is not observed. This can be explained by the way these metrics are calculated by tensorflow (27): Training-metrics are calculated on the average *of* the last epoch while validation-metrics are calculated on the average *after* the last epoch. Furthermore BatchNormalization and Dropout layers are deactivated during validation.

Due to the callback EarlyStopping training is stopped after 5 epochs with no improvement in validation-loss, but it can be observed that training and validation curves trend towards each other. This shows the risk of overfitting if training wasn't stopped or the best model wouldn't be saved.

### 3.2 Evaluation Metrics

**Confusion Matrix:** The results of TP, TN, FP and FN are similar for fold1, fold2 and fold3 as well as the two ensembles (Random Forest and Soft Majority Vote) (see table 2). However, these results should be more stable for Random Forest and Soft Majority Vote than for the individual folds, since the ensembles combine the predications of the individual folds.

**Table 2:** TP, TN, FP and FN for the 3 folds and two ensembles. The output is presented in a table instead of the original images, which are produced by the scikit-learn confusion-matrix function, to give a better overview.

|  | TP | TN | FP | FN |
|---|---|---|---|---|
| **fold1** | 188 | 193 | 8 | 14 |
| **fold2** | 188 | 195 | 6 | 14 |
| **fold3** | 190 | 190 | 11 | 12 |
| **Random Forest** | 185 | 193 | 8 | 16 |
| **Soft Majority Vote** | 188 | 193 | 8 | 14 |

**Classification Report and f1-Score:** The similarity of the true- and false- positives and negatives is also reflected in similar macro-average f1-scores (see figure 5). However, the macro-average f1-scores for the ensembles should be more stable.

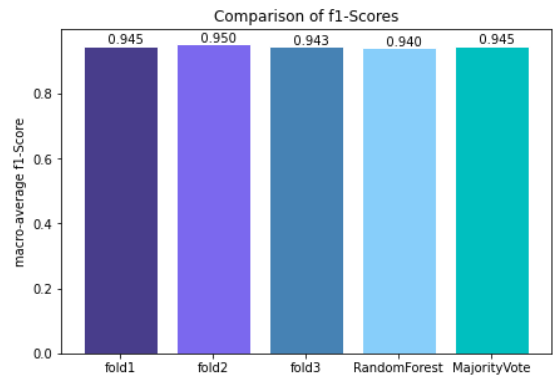

**Figure 5:** Comparison of the macro f1-Scores for the different folds and the Random-Forest and Soft Majority-Vote Ensemble.

Precision and recall are above .90 for all folds and ensembles. Recall for the viral pneumonia class, the metric with the greater importance in most medical contexts, is slightly higher for the Random Forest ensemble compared to the Soft Majority Vote ensemble, though differences might be due to chance (see table 3).

Table 3: Precision and recall for Random Forest and Soft Majority Vote ensembles for different classes.

|  | normal | | viral pneumonia | |
| --- | --- | --- | --- | --- |
|  | Random Forest | S. Majority Vote | Random Forest | S. Majority Vote |
| precision | 0.92 | 0.93 | 0.96 | 0.96 |
| recall | 0.96 | 0.96 | 0.92 | 0.93 |

**ROC-Curves:** The ROC-Curves for the different folds (see figure 6) and the ensembles (see figure 7) validate, that the classifier works well.

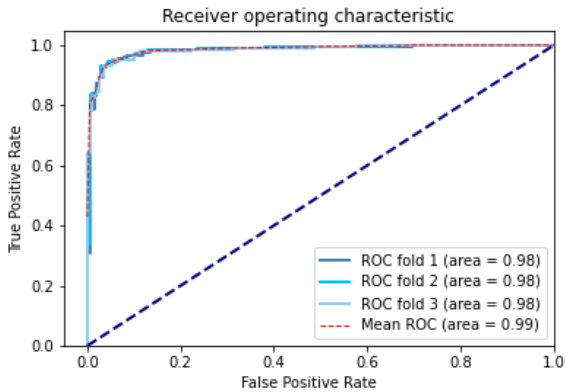

**Figure 6:** ROC-Corves for the different folds and their average.

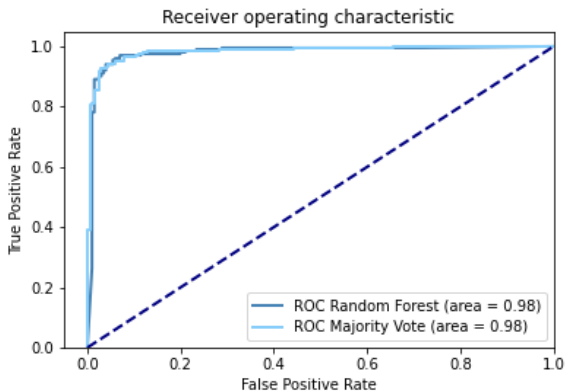

**Figure 7:** ROC-Curves for the ensembles (Random Forest and Soft Majority Vote).

## 3.5 Results of XAI

For most images, the heatmaps of Grad-CAM for the different folds are similar (see figure 8). Therefore the averaging of the three heatmaps works well (see figure 9). The output of the LIME-XAI algorithm is, in general, more difficult to compare (see figure 10).

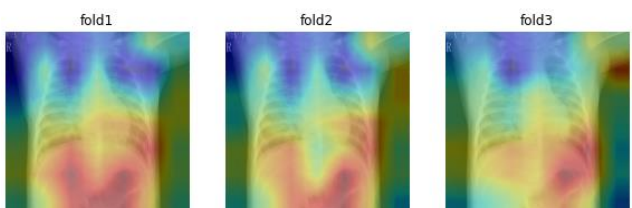

**Figure 8:** Example image showing the Grad-CAM heatmaps for the different folds (Image 195, viral pneumonia).

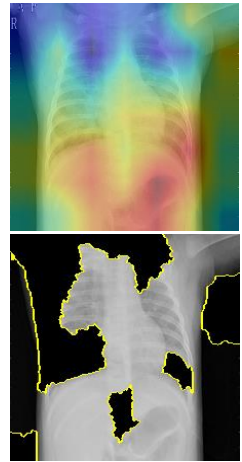

**Figure 9:** Average of the heatmaps of the 3 folds for the same image as for figure 6 (Image 195, viral pneumonia)

**Figure 10:** Output of the LIME-XAI-Algorithm for the same image as for figure 6 and 7 (Image 195, fold1, viral pneumonia)

## 4. DISCUSSION

As mentioned, the American Thoracic Society recommends, that the diagnosis of pneumonia should be made based on chest radiography. Here, X-ray images are used to differentiate "viral pneumonia" from "normal" cases. However, X-ray images can also be used to determine the source of pneumonia, bacterial or viral (2). The distinction is important for the correct choice of treatment. Therefore it would be interesting to test this deep-learning pipeline on such a task. Chest X-ray images of bacterial pneumonia are also provided by Kermany et al. (4) on kaggle (5).

Viral pneumonia affects mostly children younger than 5 years and adults older than 75 years (2). However, the image-set only contains images of children 5 years and younger. As the classifier is trained on this images-set it might not work well, when predicting on images of adults. This is a problem that commonly arises in medical contexts and points to the importance of representative training-datasets.

The metrics of the classifier show, that it seems to perform well. They should be more stable (less variance) for the ensembles then for the different folds. But empirical testing over several independent runs of the pipeline would be needed in order to evaluate the stability of the results. Moreover it would be important to compare the metrics to established methods such as PCR.

Furthermore it would be interesting to compare the output of different XAI-methods in a quantitative way to validate if the same features of the images are marked as relevant.

In this regard the interrater-reliability between medical professionals and XAI-algorithms regarding the most informative image-features would also provide further inside.

Finally, in cases where enough data is available it might be useful to spare a separate data-set or data-sets for final evaluation after the complete pipeline is setup and tested. Since the test-set is used before the final pipeline-settings are found and thus the hyperparameters might be (even if unintentionally) adjusted based on the evaluation results of the test-set there is some information-leak form the test-set into the model, which should be avoided.